\documentclass[letterpaper, 10 pt, journal, twoside]{IEEEtran}
\usepackage[bb=dsserif]{mathalpha}

\usepackage{tikz}
\usepackage{xcolor}
\usepackage{balance}
\usepackage{comment}
\usepackage{multirow}
\usepackage{graphicx}
\usepackage{booktabs}
\usepackage{stfloats}

\graphicspath{{./imgs/}}

\usepackage{amsmath}
\usepackage{amssymb}

\usepackage{tabularx}
\usepackage{wrapfig}

\usepackage{url}
\usepackage{caption}
\usepackage{subcaption}
\usepackage{cite}

\usepackage{algpseudocode}
\usepackage{algorithm}

\newcommand{\G}{\square}
\newcommand{\F}{\Diamond}
\newcommand{\R}{\mathbb{R}}
\newcommand{\W}{\mathcal{W}}
\newcommand{\mat}[1]{\begin{bmatrix} #1 \end{bmatrix}}

\newtheorem{prob}{Problem}
\newtheorem{prop}{Proposition}
\newtheorem{defn}{Definition}
\newtheorem{lemma}{Lemma}

\newtheorem{theorem}{Theorem}
\newtheorem{exmp}{Example}

\newtheorem{remark}{Remark}

\begin{document}
	\markboth{IEEE Robotics and Automation Letters. Preprint Version. Accepted February, 2024}
	{Karagulle \MakeLowercase{\textit{et al.}}: Safe Preference Learning} 
	
	\author{Ruya Karagulle$^{1}$, Nikos Ar\'echiga$^{2}$, Andrew Best$^{2}$, Jonathan DeCastro$^{2}$, and Necmiye Ozay$^{1}$ 
		\thanks{Manuscript received: September 27, 2023; Revised December 4, 2023; Accepted February 13 2024.}
		\thanks{This paper was recommended for publication by Editor Angelika Peer upon evaluation of the Associate Editor and Reviewers' comments.} 
		\thanks{Toyota Research Institute provided funds to support this work.}
		\thanks{A poster abstract on our preliminary results was presented at
			HSCC \cite{Karagulle2023}.}
		\thanks{$^{1}$Electrical and Computer Engineering, University of Michigan, Ann Arbor, USA
			{\tt\small \{ruyakrgl, necmiye\}@umich.edu}}
		\thanks{$^{2}$ Toyota Research Institute, Los Altos, USA 
			{\tt\small \{andrew.best, nikos.arechiga, jonathan.decastro\}@tri.global}}
		\thanks{Digital Object Identifier (DOI): see top of this page.}
	}
	
	\title{A Safe Preference Learning Approach for Personalization with Applications to Autonomous Vehicles}
	
	\maketitle
	
	\begin{abstract}
		This work introduces a preference learning method that ensures adherence to given specifications, with an application to autonomous vehicles. Our approach incorporates the priority ordering of Signal Temporal Logic (STL) formulas describing traffic rules into a learning framework. By leveraging Parametric Weighted Signal Temporal Logic (PWSTL), we formulate the problem of safety-guaranteed preference learning based on pairwise comparisons and propose an approach to solve this learning problem. Our approach finds a feasible valuation for the weights of the given PWSTL formula such that, with these weights, preferred signals have weighted quantitative satisfaction measures greater than their non-preferred counterparts. The feasible valuation of weights given by our approach leads to a weighted STL formula that can be used in correct-and-custom-by-construction controller synthesis. We demonstrate the performance of our method with a pilot human subject study in two different simulated driving scenarios involving a stop sign and a pedestrian crossing. Our approach yields competitive results compared to existing preference learning methods in terms of capturing preferences and notably outperforms them when safety is considered.
	\end{abstract}

	\section{Introduction}\label{sec:intro}
	\IEEEPARstart{P}{references} are a fundamental aspect of human behavior and decision-making, and it is valuable to design autonomous systems that allow for personalization to better suit the needs and desires of users. In particular for autonomous driving, surveys have demonstrated that drivers have different comfort and performance preferences while driving in different scenarios and conditions \cite{Hasenjager2017, Park2020, Bellem2018}. Moreover, drivers tend to prefer different driving styles for autonomous vehicles than their own styles \cite{Basu2017}. Customizing autonomous vehicles based on user preferences can increase user satisfaction with these vehicles. However, autonomous systems often require the satisfaction of a set of rules for safe operation. Relying on human preferences alone may result in unsafe behaviors. For instance, at an intersection with a stop sign, drivers may sometimes prefer a rolling stop, which is illegal, over a full stop. However, an autonomous vehicle should always stop completely at a stop sign to guarantee the safety of all agents in the environment. Preference learning algorithms for safety-critical operations must consider rule satisfaction. The main motivation for our work is the need for safe, trustworthy, and customizable autonomous vehicle algorithms.
	
	For safety-critical applications like driving, there are three desirable properties a preference learning method should satisfy to allow safe personalization: (i) \emph{expressivity}: the model should be expressive enough to capture preferences; (ii) \emph{safety}: it ensures safety by preferring a rule-following behavior against a rule-violating one (even in cases where the latter is scarce in the training data); and (iii) \emph{usability in control design}: the learned model should be easy to integrate into downstream correct-by-construction control synthesis tasks. In this work, we propose an integrated framework for personalization and safety to satisfy all of these properties by using Signal Temporal Logic (STL). STL is a variant of temporal logic that is tailored for reasoning about the temporal properties of time series data and is commonly used in describing correct behaviors in a variety of autonomous systems \cite{Lindemann2019, Raman2014, Kloetzer2007, Bombara2018, Li2023, Puranic2022}. 
	
	To develop a personalization framework with the STL formalism, we use a parametric extension to Weighted Signal Temporal Logic (WSTL), which is tailored for the ordering of preferences and priorities in STL formulas \cite{Mehdipour2021}. We introduce a learning framework that is based on this extension. The learning framework returns the required parameters for the WSTL formula, which can be used to synthesize a controller that yields preferred system behaviors, as in \cite{Mehdipour2021}, \cite{Cardona2023}. Starting with a parametric WSTL formula that specifies task objectives (traffic rules in autonomous vehicles) and a set of pairwise comparison preferences among a set of safe behaviors, the goal is to find suitable formula parameters such that preferred signals have greater satisfaction measure, namely \textit{WSTL robustness}, than their non-preferred counterparts. We show how to cast this problem as an optimization problem. We propose two different approaches to solve the resulting optimization problem: a random sampling approach and a gradient-based approach, which utilizes computation graphs to calculate the WSTL robustness of signals. 
	
	To evaluate the performance of our framework, we simulate two different driving scenarios: one with an autonomous vehicle navigating an intersection with a stop sign and one with an autonomous vehicle approaching a crosswalk while a pedestrian is crossing the road. We generate two sets of trajectories that comply with traffic rules for these scenarios and run a pilot human subject study with eight participants for both scenarios. Comparisons with baseline preference learning methods verify the need for safety-aware preference learning by showing that baseline methods usually lead to unsafe selections, whereas our method does not. 
	
	\section{Literature Review}\label{sec:litrev}
	
	Preference learning aims to understand and predict individuals' preferences based on a set of their choices \cite{Furnkranz2011, Martyn2023}. This can be done through independent evaluations, such as ratings or comparisons with alternatives. While both evaluations open new research methods, learning from comparison pairs may help in terms of dividing the problem into smaller, more manageable batches \cite{Furnkranz2011}. While these methods capture and reason about preferences, for safety-critical scenarios such as capturing driving preferences and personalizing driving styles, they cannot ensure necessary safety guarantees. 
	
	Another use of preferences is preference-based learning for reward functions and task learning in robot systems \cite{Sadigh2017, Biyik2018, Tucker2021}. For safety-aware applications, \cite{Cosner2022} combines preference-based learning with control barrier functions. 
	
	On the other hand, encoding safety rules in temporal logic is an eminent method for safety-critical applications \cite{Plaku2016, Nilsson2016}. Specifications in temporal logic can be used for controller synthesis \cite{Lindemann2019, Raman2014, Nilsson2016}, motion planning \cite{Kloetzer2007, Fainekos2009, Linard23} and learning applications \cite{Neider2019, Bombara2018, Li2023, Xu2019, Chou2020, Puranic2022, Jiang2021, Li2018} in many autonomous systems. In particular, works in \cite{Neider2019, Bombara2018, Li2023, Xu2019, Karagulle2022} try to infer a temporal logic formula for classification from the data. As a subset of learning applications, in robot learning, Chou et al. \cite{Chou2020} try to learn task specifications in linear temporal logic from demonstrations, Puranic et al. \cite{Puranic2022} score demonstrations with the help of ordered specifications in the form of signal temporal logic and works in \cite{Jiang2021, Li2018} use temporal logic for reward shaping and reinforcement learning. 
	
	Incorporating preferences and priorities with temporal logic is studied in \cite{Mehdipour2021, Wang2022, Yan2021, Fronda2022, Karagulle2023}. The work in \cite{Mehdipour2021} introduces a weighted variant of the STL, called Weighted Signal Temporal Logic (WSTL), in which weights reflect the order of priorities or preferences. The work in \cite{Wang2022} defines Weighted Truncated Linear Temporal Logic. Both works assume that they have knowledge of the formula and associated weights. For the end-user, it is hard to interpret the weights and define their preferences in the temporal logic formalism, so there needs to be an intermediate step to infer the weights from the user. In \cite{Yan2021, Fronda2022}, a parametric extension of WSTL, which we call PWSTL, is used in a time series classification problem, where weights of the formula are learned using neural networks. 
	
	\section{Preliminaries}
	\subsection{Signal Temporal Logic (STL)}
	STL is a temporal logic formalism used to reason about signals $s: \mathbb{T} \to \mathcal{S}$, where $\mathbb{T}$ is a time domain and $\mathcal{S} \subseteq \R^{m}_e$ is a $m$ dimensional extended real-valued signal domain \cite{Maler2004}. We will consider $\mathbb{T}$ to be infinite $\mathbb{Z}_{\geq 0}$ or finite $[0, t_{final}]\subset \mathbb{Z}_{\geq 0}$.
	An STL formula $\phi$ is given by the grammar $\phi ::= \top \mid \pi \mid \lnot \phi \mid \phi_1 \wedge \phi_2 \mid \phi_1 \mathcal{U}_{[a,b]} \phi_2.$ Boolean true is $\top$, and $\pi$ is a predicate of the form $\pi(s(t)):= f_\pi(s(t))\geq 0$ where $f_\pi: \mathcal{S} \to \R_e$ and $s(t)$ is the signal value at time instant $t$. The logical not is $\lnot$, the conjunction is $\wedge$, and $\mathcal{U}_{[a,b]}$ is the ``Until" operator. Additional operators, disjunction $\vee$, Always $\G_{[a,b]}$, and Eventually $\F_{[a,b]}$ can be derived from operators in the grammar\footnote{Disjunction is $\phi_1\vee \phi_2 = \lnot(\lnot \phi_1 \wedge \lnot \phi_2)$, Eventually is $\F_{[a,b]}\phi = \top \mathcal{U}_{[a,b]} \phi$, and Always is $\G_{[a,b]}\phi = \lnot(\F_{[a,b]}\lnot \phi)$.}. Subscript ${[a,b]}$ defines the time interval. When the time interval is from $0$ to $\infty$, the subscript is omitted. We will denote the set of all well-formed STL formulas with $\cal{F}$. If a signal $s$ satisfies a formula $\phi$ at time $t$, it is shown as $(s,t) \models \phi$. If it violates at $t$, it is shown as $(s,t) \not \models \phi$. The qualitative semantics of STL are defined as follows:
	\begin{equation*}\arraycolsep=2pt
		\begin{array}{lll}
			(s,t) \models \pi & \Leftrightarrow & \pi(s(t)), \\
			(s,t) \models \lnot \phi & \Leftrightarrow &  (s,t) \not \models \phi,\\
			(s,t) \models \phi_1 \wedge \phi_2 &\Leftrightarrow& ((s,t) \models \phi_1 \text{ and } (s,t) \models \phi_2),\\
			(s,t) \models \phi_1 \mathcal{U}_{[a,b]} \phi_2 & \Leftrightarrow &\exists t' \in [t+a, t+b] ((s, t') \models \phi_2 \\ & & \text{and } \forall t'' \in [t, t') \hspace{0.25em} (s, t'') \models \phi_1  ).
		\end{array}
	\end{equation*}
	Derived operators have the following qualitative semantics: 
	\begin{equation*}
		\begin{array}{lll}
			(s,t) \models \phi_1 \vee \phi_2 & \Leftrightarrow &((s,t) \models \phi_1 \text{ or } (s,t) \models \phi_2), \\
			(s,t) \models \G_{[a,b]} \phi & \Leftrightarrow& \forall t' \in [t+a, t+b] \hspace{0.25em} (s,t') \models \phi,\\
			(s,t) \models \F_{[a,b]} \phi & \Leftrightarrow &\exists t' \in [t+a,t+b]   \hspace{0.25em} (s, t') \models \phi. \\ 
		\end{array}
	\end{equation*}
	
	For qualitative semantics at time instant $t=0$, we omit $t$ and write $s \models \phi$. STL also has quantitative semantics to measure how well the signal models the formula. There are different quantitative semantics, also known as robustness metrics \cite{Donze2010, Varnai2020}. In this paper, we use the traditional robustness metric $\rho: \mathcal{S} \times \mathcal{F} \times \mathbb{T} \to \R_e$ from \cite{Donze2010}, defined recursively as\footnote{To represent Boolean-valued quantities, we use signals $p:\mathbb{T}\to \{-\infty,+\infty\}$ and simply write $p$ as a predicate instead of $p\geq 0$. If such a signal is the $i^{th}$ coordinate of $s$ and $\pi(s(t)) := e_i^Ts(t)\geq 0$ with $e_i$ being the $i^{th}$ natural basis vector, we have $\rho(s,\pi,t) = e_i^Ts(t) = p(t)$.}:	
	\begin{equation*}\arraycolsep=2pt
		\begin{array}{rcl}
			\rho(s, \top, t) &=& \infty, \\
			\rho(s, \pi, t) &=& f_\pi(s(t)), \\
			\rho(s, \lnot \phi, t) &=& -\rho(s,\phi, t), \\
			\rho(s, \phi_1 \wedge \phi_2, t) &=& \min \big (\rho(s, \phi_1, t), \rho(s,\phi_2,t)\big ), \\
			\rho(s, \phi_1 \mathcal{U}_{[a,b]} \phi_2, t ) &=& \max\limits_{t' \in [t+a,t+b]}\big( \min \big(\rho(s, \phi_2, t'), \\ & &\hspace{3.5em}\min\limits_{t'' \in [t,t']}\rho(s,\phi_1,t'')\big )\big ).
		\end{array}
	\end{equation*}
	The robustness for derived operators is given by 
	\begin{equation*}\arraycolsep=2pt
		\begin{array}{rcl}
			\rho(s, \phi_1 \vee \phi_2, t) &=& \max \big (\rho(s, \phi_1, t), \rho(s,\phi_2,t)\big ), \\
			\rho(s, \Diamond_{[a,b]} \phi, t ) &=& \max\limits_{t' \in [t+a, t+b]}\rho(s, \phi, t'), \\
			\rho(s, \square_{[a,b]} \phi, t ) &=& \min\limits_{t' \in [t+a, t+b]} \rho(s, \phi, t'). \\
		\end{array}
	\end{equation*}
	The robustness value at $t=0$ is shown as $\rho(s, \phi)$. Note that for finite signals where $t_{final} < \infty$, time interval $[t+a, t+b]$ in temporal operators may exceed the time length of the signal. In this case, time interval can be taken as $[t+a, \min(t+b, t_{final})]$ assuming that $t+a \leq t_{final}$.
	For simplicity, we keep the semantics for infinite signals but we use STL for finite signals with necessary corrections\cite{DeGiacomo2013}. The robustness metric $\rho$ is sound, i.e., $\rho(s, \phi, t) > 0 \implies (s,t) \models \phi$ and $\rho(s, \phi, t) < 0 \implies (s,t) \not\models \phi$ \cite{linearuntil2013}.
	
	\begin{exmp}\label{exp:1}
		Let $s = \mat{s_1 & s_2}^{T} = \begin{bmatrix}1 & -1 & -2 & -2 \\ 1 & 1 & 1 & 2  \end{bmatrix} \in \R^{2 \times 4}$ be a two-dimensional signal with the time length $t_{final} =3$. Let $\phi_{\rm STL} = \F( -s_1 \geq 0 \wedge s_2 \geq 0 )$ be an STL formula. Satisfaction of $\phi$ by the signal $s$ means that ``There is a time $t^* \leq t_{final}$ such that $s_1(t^*) \leq 0$ and $s_2(t^*) \geq 0$". The robustness of $s$ at time $t=0$ over $\phi_{\rm STL}$ is \[\rho(s, \phi_{\rm STL}) = \max_{t'\in [0,3]}( \min( -s_1(t'), s_2(t') ) ) =  2.\] 
	\end{exmp}

	\subsection{Weighted Signal Temporal Logic (WSTL)} \label{sec:wstl}
	WSTL is tailored to represent priorities and preferences in STL formulas \cite{Mehdipour2021}. Its syntax extends STL syntax as $$ \phi := \top \mid \pi \mid \lnot \phi \mid \phi_{1} \wedge^{w} \phi_{2} \mid \phi_{1} \mathcal{U}_{[a,b]}^{w^1,w^2} \phi_{2},$$
	where the weights are $w\in\mathbb{R}_+^2$ and $w^1,w^2\in\mathbb{R}_+^{(b-a+1)}$. All operators are interpreted as in STL.
	
	In \cite{Mehdipour2021}, the quantitative semantics of WSTL is called \textit{the WSTL robustness}, denoted as $r: \mathcal{S} \times \mathcal{F} \times \mathbb{T} \to \R_e$. We adopt the WSTL formalism with the following quantitative semantics:
	\begin{equation}\label{eq:quantsemantics}
		\arraycolsep=1pt
		\begin{array}{rcl}
			r(s, \top, t) &=& \infty \\
			r(s, \pi, t) &=& \rho(s,\pi,t) \\
			r(s, \lnot \phi, t) &=& -r(s,\phi, t), \\
			r(s, \phi_1 \wedge^{w} \phi_2, t) &=& \min \big (w_1r(s, \phi_{1},t),  w_2r(s, \phi_{2},t) \big ), \\
			r(s, \phi_1 \mathcal{U}_{[a,b]}^{w^1,w^2} \phi_2, t ) \hspace{-.2em}  &=& \hspace{-.8em}  \max\limits_{t' \in [t+a,t+b]} \hspace{-.4em}  \Big( \hspace{-.3em}  \min \hspace{-.3em} \big(  w^1_{t'-t-a+1}r(s, \phi_2, t'), \\  &&\hspace{1.5em}   w^2_{t'-t-a+1} \min\limits_{t'' \in [t,t')}r(s,\phi_1,t'')\big )\hspace{-.3em} \Big ).
		\end{array}
	\end{equation}
	Derived operators have WSTL robustness definitions as:
	\begin{equation*}\label{eq:quantsemantics_derived}
		\arraycolsep=2pt
		\begin{array}{rcl}
			r(s, \phi_{1} \vee^{w} \phi_2, t) &=& \max \big(w_1r(s, \phi_{1},t), w_2r(s, \phi_{2},t)\big), \\
			r(s, \G_{[a,b]}^{w} \phi, t ) &=& \min\limits_{t' \in [t+a, t+b]}(w_{t'-t-a+1}r(s, \phi,t')), \\
			r(s, \F_{[a,b]}^{w} \phi, t ) &=& \max\limits_{ t' \in [t+a, t+b]}(w_{t'-t-a+1}r(s, \phi,t')),
		\end{array}
	\end{equation*}
	with $r(s, \phi)$ denoting the WSTL robustness at $t=0$.
	
	Note that since we have $\top$ when defining Eventually from Until, and since the WSTL robustness of $\top$ is $\infty$, we drop the set of weights $w^2$ in the WSTL robustness of Eventually because they do not affect the result of the \texttt{min} operation in the computation of the WSTL robustness of Until. That is the reason why Eventually (and hence Always) has fewer weights than Until. Moreover, the Boolean true, predicates, and negation operators do not have associated weights, i.e., these operators have weights equal to $1$.
	
	\addtocounter{exmp}{-1}
	\begin{exmp}(cont.)
		Let $\phi$ be the weighted version of $\phi_{\rm STL}$ with weights $\{w_i^\F\}_{i=1}^{4} =  [1.5, 0.3, 3, 1.2]$ and $\{w^\wedge_i\}_{i=1}^2 = [1,2]$.
		The WSTL robustness of $s$ at time $t=0$ is 
		\[r(s, \phi) = \max_{t' \in [0,3]}( w^{\F}_{t'+1}\min( -w^{\wedge}_{1}s_1(t'), w^{\wedge}_2s_2(t') ) ) = 6.\]
	\end{exmp}
	
	The following result is adapted from  Theorem 2 in \cite{Mehdipour2021}.
	\begin{lemma} \label{th:WSTLsoundness}
		Let $\tilde{r}: \mathcal{S} \times \mathcal{F} \times \mathbb{T} \to \R$ be a quantitative semantics. For a WSTL formula $\phi$, let $\phi_{\rm STL}$ be the STL formula obtained by removing the weights in $\phi$.  If $\text{sign}(\rho(s,\phi_{\rm STL},t)) = \text{sign}(\tilde{r}(s,\phi,t))$ for all $(s, \phi, t) \in \mathcal{S}\times\mathcal{F} \times \mathbb{T}$  (i.e., $\tilde{r}$ is \textit{sign-consistent}), then $\tilde{r}$ is sound.
	\end{lemma}
	
	\begin{theorem}
		Quantitative semantics in \eqref{eq:quantsemantics} is sound.
	\end{theorem}
	\begin{IEEEproof}
		According to Lemma~\ref{th:WSTLsoundness}, it is sufficient to prove that quantitative semantics in \eqref{eq:quantsemantics} is sign-consistent. Since all weights are defined as positive, multiplying a robustness value with a weight does not change its sign. Therefore, for each recursive operation in the WSTL robustness calculation, the sign of the robustness value associated with this recursion step is preserved. Then, we see that $\text{sign}(\rho(s,\phi_{\rm STL},t)) = \text{sign}(\tilde{r}(s,\phi,t))$ for all $(s, \phi, t) \in \mathcal{S}\times\mathcal{F} \times \mathbb{T}$ and for all non-negative weights.
	\end{IEEEproof}
	
	In the WSTL definition of \cite{Mehdipour2021}, weights are pre-determined positive real values. In this work, we use an extension to WSTL that we call Parametric Weighted Signal Temporal Logic (PWSTL) in which some of the weights are unknown parameters and the remaining weights are given constants (cf., \cite{Yan2021}). We denote the set of unknown parameters as $\W$ and denote PWSTL formulas as $\phi_{\W}$, where we omit the known weights with slight abuse of notation since for most of the results in the letter, $\W$ is the entire weight set. A PWSTL formula results in a WSTL formula $\phi_{\W=w}$ with the valuation $w$ of the parameters.

	\section{Problem Statement and Solution Method}
	As we focus on driving scenarios, inputs to our problem are signals. Preferences are given in pairs and preference data for signals is defined as follows.
	\begin{defn}[Preference Data]
		Preference data $\mathcal{P}:= \{(s_i^+, s_i^-)\}_{i=1}^P$ is a set of $P$ pairwise comparisons. In each pair $(s_i^+, s_i^-)$, $s_i^+$ represents the preferred signal and $s_i^-$ represents non-preferred one.
	\end{defn}
	
	The goal of this work is to select a weight valuation $\tilde{w}$ for the parameter set $\W$ of the PWSTL formula $\phi_{\W}$. Formula $\phi_{\W}$ is determined according to system rules so that it reflects safety specifications. Formally, this paper aims to solve the following problem:
	\begin{prob}\label{prob:prob}
		Given a PWSTL formula $\phi_\W$ with a weight parameter set $\W$, and a preference data $\mathcal{P}$, find a valuation $w$ of $\W$ such that 
		\begin{equation}\label{eq:ineq}
			r(s_i^+, \phi_{\W=w}) > r(s_i^-, \phi_{\W=w}) \quad \forall (s_i^+, s_i^-) \in \mathcal{P}.
		\end{equation}
	\end{prob}
	
\begin{figure}
		\centering
		\includegraphics[width= \columnwidth]{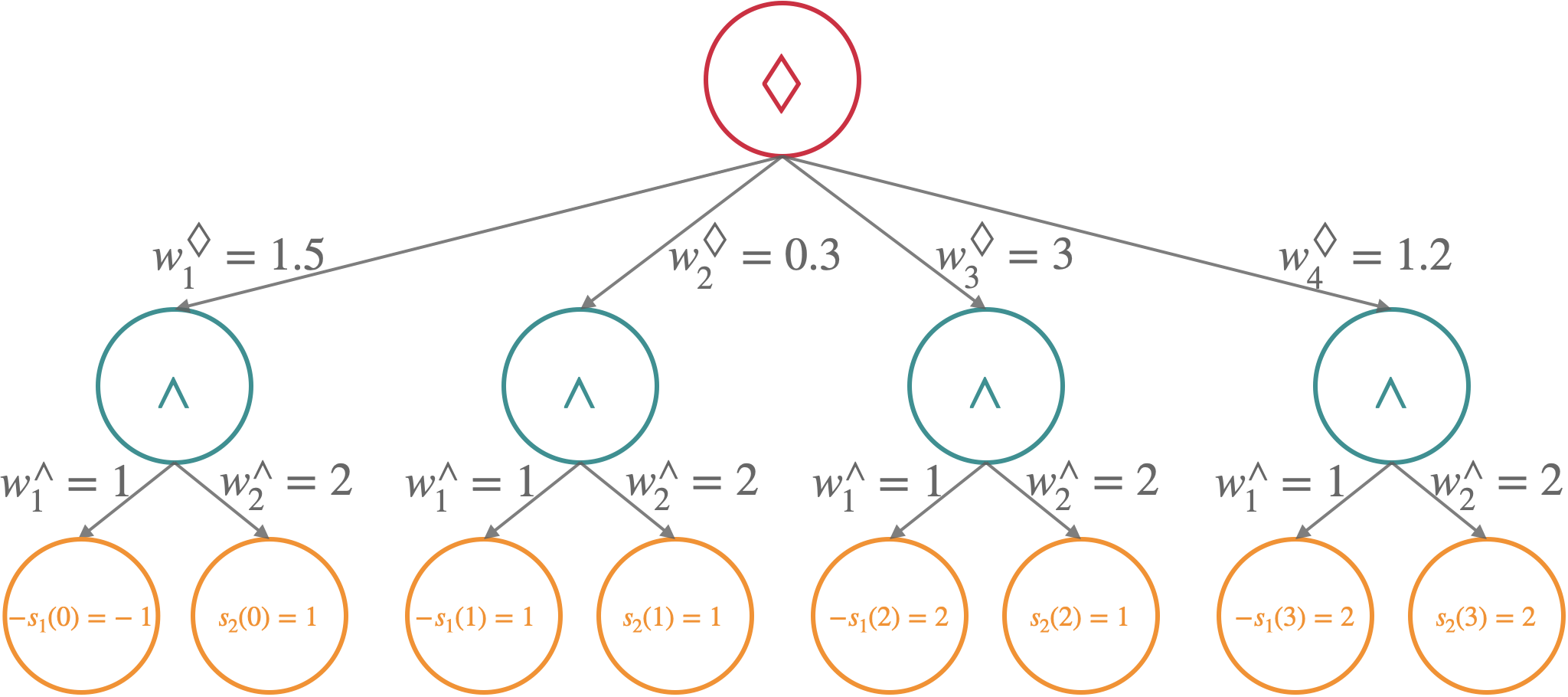}
		\caption{Syntax tree of $\phi$ of Example~\ref{exp:1}.}
		\label{fig:syntax_tree}
	\end{figure}

	Problem \ref{prob:prob} is a feasibility problem. In the next subsection, we reformulate it as an optimization problem to be able to handle infeasibility. Before doing so, we provide an analysis of the set of feasible weights using the syntax tree of STL formulas. An STL formula has an associated syntax tree, in which nodes represent Boolean and temporal operators, leaf nodes represent predicates, and edges represent the connection between operators and operands \cite{Li2021}. The syntax tree associated with the STL formula in Example~\ref{exp:1} is given in Figure~\ref{fig:syntax_tree}. Let the \emph{root weights} of a WSTL formula be the weights associated with the weighted operator closest to the root of its syntax tree. For instance, for $\phi$ of Example~\ref{exp:1}, the root of the syntax tree is $\F$ operator and the root weights are $\{w_i^\Diamond\}_{i=1}^{4}$. Consider another example with $\tilde \phi = \lnot \phi$, although the root operator is $\lnot$ since it is not a weighted operator, we need to look at its children until we find a weighted operator. Hence, the root weights are again $\{w_i^\Diamond\}_{i=1}^{4}$.
	
	Next, we show that the feasible weight valuations of Problem~\ref{prob:prob} are unbounded, when non-empty, due to homogeneity with respect to the root weights of the formula.
	
	\begin{lemma}\label{lemma:scale}
		Let $\phi_\W$ be a PWSTL formula with weight set $\W $ containing only the weight parameters for the root weights of $\phi$. Other weights of $\phi$ are fixed constants. If valuation $w$ of $\W$ solves Problem~\ref{prob:prob}, then $\tilde{w} = \alpha w$ also solves the problem for any $\alpha >0$.
	\end{lemma}
	\begin{IEEEproof}
		If the WSTL formula with valuation $w$ is a feasible solution for Problem~\ref{prob:prob}, we know that for all pairs in $\mathcal{P}$, $r(s_i^+, \phi_{\W=w}) > r(s_i^-, \phi_{\W=w})$ holds. We also have 
		\[r(s, \phi_{\W=\tilde{w}}) = \alpha r(s, \phi_{\W = w}).\]
		This together with $\alpha>0$ implies for all $(s_i^+, s_i^-) \in \mathcal{P}$, $r(s_i^+, \phi_{\W=\tilde{w}}) > r(s_i^-, \phi_{\W=\tilde{w}})$. Hence, the WSTL formula with valuation $\tilde{w}$ is a feasible solution for Problem~\ref{prob:prob}.
	\end{IEEEproof}
	
	Given the above property, namely \textit{root-level homogeneity}, we will show that it is possible to restrict the weight valuations to a bounded set $\mathcal{D}$ that is guaranteed to include at least one solution whenever a solution exists.
	
	\begin{theorem}\label{th:bounded}
		Let $\mathcal{D} = \mathcal{B}_\infty(0) \cap\mathbb{R}_+^n$, i.e., the intersection of the $n$-dimensional closed unit ball in infinity-norm and the positive quadrant. If Problem~\ref{prob:prob} is feasible with weight valuation $w$,  then there exists at least one weight valuation $\bar w$ in the domain $\mathcal{D}$ such that $\phi_{\W = \bar w}$ solves the problem.
	\end{theorem}
	\begin{IEEEproof} 
		Let Problem~\ref{prob:prob} be feasible for the valuation $w$. If $w\in\mathcal{D}$, the proof is trivial. So, let us assume $w\notin\mathcal{D}$. 
		
		We will prove the theorem by induction on the depth $d$ of the syntax tree of $\phi_{\W = w}$. For each subformula $\phi_{s}$ at level $k$ ($k<d$) of the syntax tree of $\phi_{\W = w}$, assume that the root weights of $\phi_{s}$ are $w_s$ and all the remaining weights of $\phi_{s}$ are already less than or equal to $1$ (note that this trivially holds in the base case when $k=d$ where we pick $w^{(d)} = w$). Then, we will show that we can define a new set of weights $w^{(k)}$ for $\phi_{\W}$ such that $r(s,\phi_{\W = w^{(k)}},t) = r(s,\phi_{\W = w^{(k+1)}},t)$ such that the weights of each subformula at level $k-1$ except for their root weights are less than or equal to $1$. 
		
		Consider an arbitrary subformula $\phi_{s}$ at level $k$ with weights $w_s$ satisfying the induction hypothesis. We use $\phi_{s, w_s}$ as a shorthand for such a pair to differentiate it from the same formula with updated weights, $\phi_{s, \bar w_s}$. Define $\bar w_s = w_s/\max{(w_s)}$. Clearly, $r(s, \phi_{s, w_s}, t) = \max{(w_s)}r(s, \phi_{s, \bar w_s},t )$ and all weights of $\phi_{s, \bar w_s}$ are less than or equal to $1$. However, we can scale the weights $w_u$ that multiply $r(s, \phi_{s, w_s}, t)$ at level $k-1$ with $\max{(w_s)}$ so that with valuation $w^{(k)}$, where the weights $\max{(w_s)}w_u$ and $\bar w_s$ are replaced by $w_u$ and $w_s$, we achieve the same WSTL robustness value, establishing the induction hypothesis.
		
		Finally, we can decrement $k$ until we reach the root weights of $\phi_{\W = w}$ and invoke Lemma~\ref{lemma:scale} to scale the root weights to be less than or equal to $1$ while preserving feasibility. Therefore, the scaled valuation is in $\mathcal{D}$.\end{IEEEproof}
	
	We illustrate the proof with our running example.
	\addtocounter{exmp}{-1}
	\begin{exmp}(cont.) The weight values $w$ of $\phi$ are not in $\mathcal{D}$. We can construct a new weight valuation that preserves preference orders using Theorem~\ref{th:bounded}. Let us denote $\phi$ as  $\phi_{\W=w}$ and consider two signals $x$ and $y$ with robustness order $r(x,\phi_{\W=w}) > r(y, \phi_{\W=w})$.
		At level $k=2$, we have $\max_i(w_i^\wedge)=2$. Define $\tilde w^\wedge = w^\wedge/\max_i(w_i^\wedge) = [0.5,1]$. Then, scale the weights in the upper level with $\max_i(w_i^\wedge)$ and obtain $\tilde w^\F =[3, 0.6, 6, 2.4]$. Note that $r(x,\phi_{\W=w}) = r(x,\phi_{\W=\tilde w})$ and $r(y,\phi_{\W=w}) = r(y,\phi_{\W=\tilde w})$. Now, let $k=1$, which is the root level, and consider $\W=\tilde w$. Leave the lower levels as is: $\bar w^{\wedge} = \tilde w^{\wedge}$ and scale the root weights as $\bar w^{\F} = \tilde w^{\F}/ \max_i(\tilde w_i^{\F}) = [0.5, 0.1, 1, 0.4]$. By root-level homogeneity, we know that, if $r(x,\phi_{\W=\tilde w}) > r(y, \phi_{\W= \tilde w})$, which is the case by construction of $\tilde w$, then $r(x, \phi_{\W=\bar w})> r(y, \phi_{\W=\bar w})$. Hence, $\bar w$ preserves the orders and it is in $\mathcal{D}$.
	\end{exmp} 
	
	Having a bounded feasible domain will be useful in our computational approach.
	\subsection{An Optimization Reformulation}
	Problem~\ref{prob:prob} can be formulated as an optimization problem. 
	\begin{prob}\label{prob:minprob}
		Given preference data $\mathcal{P}$, PWSTL formula $\phi_{\W}$ and domain $\mathcal{D}$ described in Theorem \ref{th:bounded}, solve
		\begin{equation}\label{eq:argmin}
			w^* \in \arg \min_{w \in \mathcal{D}}\hspace{-0.1in}\sum_{(s_i^+, s_i^-) \in \mathcal{P}} \hspace{-0.2in} -\mathbb{1}(w)_{(r(s_i^+, \phi_{\W=w})-r(s_i^-, \phi_{\W=w})> 0)},
		\end{equation}
		where $\mathbb{1}(w)$ is the indicator function which takes $\mathbb{1}(w) = 1$ when the subscripted condition is satisfied and takes $\mathbb{1}(w) = 0$ otherwise.\footnote{Since the objective function takes only finitely many values, it always has a minimum. Therefore, searching for $\arg \min$ is valid.} 
	\end{prob}
	
	By construction of the problem \eqref{eq:argmin}, we have the following result that states that relates the solution of this optimization problem to Problem~\ref{prob:prob}.
	\begin{prop}
		\label{th:equivalence}
		If Problem~\ref{prob:prob} is feasible, then a minimizer $w^*$  of Problem~\ref{prob:minprob} is a solution to Problem~\ref{prob:prob}. Moreover, if Problem~\ref{prob:prob} is infeasible, Problem~\ref{prob:minprob} finds a valuation for $\phi_{\mathcal{W}}$ that maximizes the number of pairs that satisfy Inequality~\eqref{eq:ineq}. 
	\end{prop}
	
	Problem~\ref{prob:minprob} not only transforms the feasibility Problem~\ref{prob:prob} into an optimization problem but also returns a valuation that makes maximum number of pairs correctly ordered according to Inequality~\eqref{eq:ineq} when Problem~\ref{prob:prob} is infeasible. 
	
	It is important to note that with Problem~\ref{prob:minprob} and weights being positive, it is impossible to find weight valuations that result in a greater robustness value of a rule-violating behavior than the robustness value of a rule-satisfying one. Violating signals will always have negative robustness values. If there exists a pair in the preference dataset such that the person prefers a rule-violating behavior over a satisfying behavior, we cannot satisfy Inequality~\eqref{eq:ineq} for this pair, Problem~\ref{prob:prob} becomes infeasible and we will find a valuation that satisfies Inequality~\eqref{eq:ineq} for maximum number of pairs. 
	
	\begin{remark}\label{rem:boolean} We note that for certain STL formulas $\phi_{\rm STL}$, the corresponding WSTL robustness metric satisfies $r(s,\phi_{\W = w}) \leq 0$ for all signals $s$ and all weights $w\in \mathbb{R}_+^n$. In this case, the problem of learning weights only from rule-satisfying signals is not meaningful since these signals will have $r(s,\phi_{\W = w}) =0$ for any weight $w$; and satisfaction cannot be deduced when $r(s,\phi_{\W = w}) =0$.
		While such formulas can be uncommon in certain domains, we find them to be common in driving scenarios when the specification involves coming to a full stop or a Boolean indicator for pedestrians or traffic lights.  Hence, we discuss a workaround to enable preference learning among satisfying signals for a class of such formulas but similar workarounds can be devised for other cases too. Consider a formula of the form $\phi = \phi_1 \wedge \G(f(s)\geq 0\; \wedge\; -f(s)\geq 0)$, second part of which essentially represents an equality $f(s)=0$. To enable preference learning in this case, we append $s$ with a new coordinate $b$ such that $b(t)=\infty$ if $f(s(t))=0$ and $b(t)=-\infty$ otherwise; and replace the formula with $\phi' = \phi_1 \wedge \G b$. With this transformation, if for all $t \in [0, t_{final}]$, $b(t)=\infty$, the WSTL robustness of $\phi'$ is determined by $\phi_1$; otherwise the WSTL robustness of $\phi'$ becomes $-\infty$, which indicates a violation of $\phi$. Applying our method to $\phi'$ allows us to rank rule-satisfying signals.   
	\end{remark}
	
	\subsection{Computational Approach}\label{sec:comp_app}
	We note that Problem~\ref{prob:minprob} is highly non-convex and non-differentiable. As a result, it is hard to solve this problem to a global minimum. In the following, we propose two approaches, one gradient-based, the other sampling-based that aim to find an approximate solution. 
	
	\paragraph{Gradient-based optimization} Thanks to the
	prevalence and success of gradient-based methods and back-propagation in machine learning, many temporal logic learning algorithms using gradients have been proposed \cite{Leung2021}. To be able to compute the gradient, we need a differentiable loss function. In the WSTL robustness definition, we replace \texttt{max} and \texttt{min} functions with their soft differentiable versions \texttt{softmin/softmax} as in \cite{Leung2021}. We also replace the indicator function with the logistic function with a shift. The shift helps avoid equality of robustness values in preference pairs. Overall, we propose the following surrogate loss 
	\begin{equation*}
		\arraycolsep=0.5pt
		\begin{array}{lr}
			\mathcal{L} = \hspace{-1.2em} \sum\limits_{(s_i^+, s_i^-) \in \mathcal{P}} \hspace{-1.3em} & \hspace{-1.2em}(1+\exp(M[r(s_i^+, \phi_{\W=w})\hspace{-.3em}- \hspace{-.3em}r(s_i^-, \phi_{\W=w})\hspace{-.3em}- \hspace{-.3em}\epsilon]))^{-1}  \\
			& + \log(1+\theta\exp(\| W_\phi \|_2^2 \hspace{-.3em}- \hspace{-.3em} \| W_\phi^{init} \|_2^2)),
		\end{array}
	\end{equation*}
	where $M$ is a large number, $\epsilon$ is a small shift, and $\theta$ is an optimization weight for the second term. Here, the first term is an approximation of the cost function in Equation~\eqref{eq:argmin} and the second term promotes the norm of the weights $W_\phi$ not to change too much compared to its initial value $W_\phi^{init}$, where $W_\phi^{init}\in \mathcal{D}$. This second term is essentially a surrogate for the constraints in Equation~\eqref{eq:argmin}; and due to Theorem~\ref{th:bounded} and the equivalence of the infinity norm and 2-norm in finite dimensions, does not change the validity of the solutions.
	
	\begin{figure*}[!hbt]
		\begin{subfigure}[t]{0.49\textwidth}
			\centering
			\includegraphics[width=1.02\textwidth]{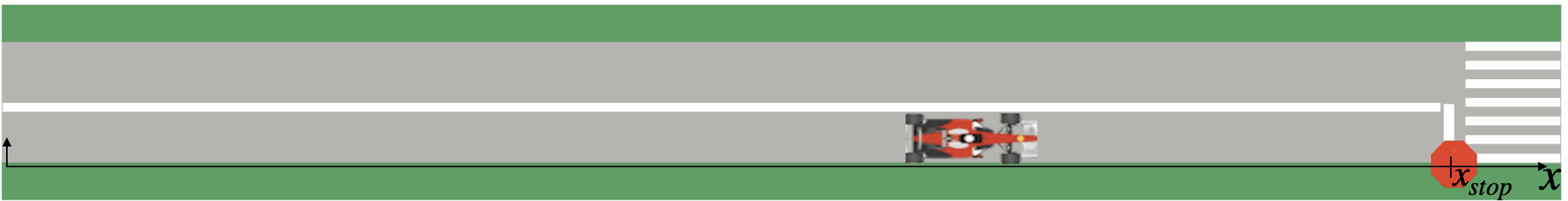}
			\caption{Stop Sign Scenario: Vehicle approaching to an intersection with a stop sign. The traffic rule says that vehicles should stop before the stop sign.}
			\label{fig:stop_sign_fig}
		\end{subfigure}
		\begin{subfigure}[t]{0.49\textwidth}
			\centering
			\includegraphics[width=.95\textwidth]{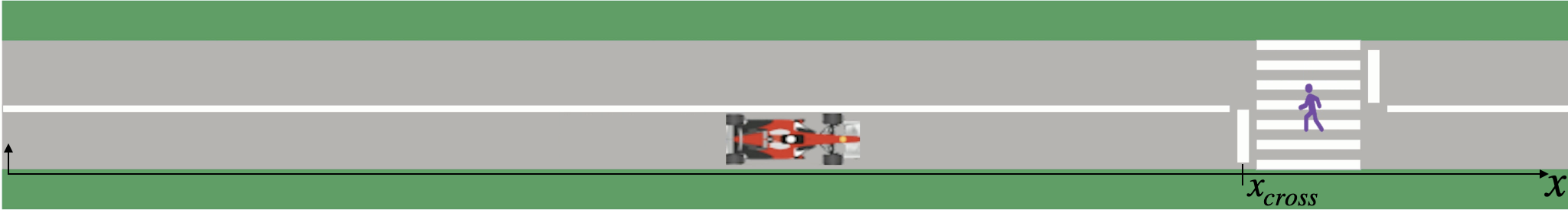}
			\caption{Pedestrian Scenario: Vehicle approaching to a pedestrian crosswalk, while a pedestrian is crossing. The vehicle can come to a complete stop or slow down sufficiently to allow the pedestrian.}
			\label{fig:pedestrian_fig}
		\end{subfigure}
		\caption{Two scenarios that are used for experiments}
	\end{figure*}
	
	\noindent\emph{Implementation details:} 
	Inspired by \cite{Leung2021}, we construct a computation graph for the robustness of WSTL formulas from syntax trees. This computation graph takes a signal as input and returns the WSTL robustness value of that signal at all times as output. We use PyTorch along with Adam \cite{Kingma2014} optimizer. Several strategies are investigated to mitigate convergence issues of the gradient-based method: (i) decreasing the softness coefficient $\beta$ of \texttt{softmin/max}, possibly compromising the soundness guarantee, (ii) decreasing the steepness of the logistic function, i.e., decreasing $M$, but this makes the surrogate $\mathcal{L}$ less similar to the objective in Problem~\ref{prob:minprob}, (iii)  initializing the iteration from multiple random points to overcome bad local minima.
	
	\paragraph{Random Sampling} Randomized methods have shown some success in temporal logic planning \cite{Kantaros2018}, especially when there is a multitude of feasible solutions. Similarly in \cite{Mania2018}, it is shown that simple random search can give not only competitive but also faster results compared to gradient methods. This inspires our attempt to solve Problem \ref{prob:minprob} through random sampling in the region $\mathcal{D} = \mathcal{B}(0)_\infty \cap \mathbb{R}_+^n$. We uniformly sample weight valuations in $\mathcal{D}$. 
	
	\noindent\emph{Implementation details:} We want weight valuations such that the absolute difference in robustness of signals within a pair should exceed $5\%$ of the range between the maximum and minimum robustness values among all signals. While this condition is not required for the random sampling approach alone, it can be useful for two downstream tasks: (i) when using the best performing of these weights as initialization of gradient-based approaches\footnote{We tried this combination in our experiments, however, the performance improvement was not significant. Therefore, due to space constraints, we do not report these results further.}, this separation helps start the iterations at a part of the weight space where the logistic function well-approximates the indicator function; (ii) when using the learned formula in controller synthesis, weights that well-separates the preferences lead to controllers that more robustly reflect the preferences. 
	
	\section{Experiments}
	In this section, we provide a comparison of solution approaches with baseline methods, along with demonstrating the need for a safety-guaranteed preference learning framework.\footnote{The code and the data can be accessed from \url{https://github.com/ruyakrgl/SPL-WSTL}} We also showcase the framework's performance in capturing the personal preferences of different participants in a human subject study. For these purposes, we use two different driving scenarios. 
	
	\emph{Driving Scenarios:} We use STL to specify traffic rules in driving scenarios. The first scenario is a simple intersection with a stop sign, a screenshot is shown in Figure~\ref{fig:stop_sign_fig}. The vehicle must stop before the stop sign, but there is some flexibility in the approach and final position. The traffic rule can be expressed as follows: $\phi^{stop} = \F\G ( x-x_{stop} \geq 0 \wedge v = 0) \wedge \G(v \geq 0)$ where $x$ and $v$ are the position and speed signals of a vehicle, respectively, and $x_{stop}$ is the stop sign position. Note that $\rho(s,\phi^{stop}) \leq 0$ for any signal due to equality condition. We substitute $v=0$ with an indicator variable as discussed in Remark~\ref{rem:boolean}. We construct the PWSTL formula $\phi^{stop}_{\W}$ with a weight parameter set $\W$ that contains all weights in the formula. In the second scenario, we observe an ego vehicle approaching a pedestrian while she is crossing the road, as illustrated in Figure~\ref{fig:pedestrian_fig}. The traffic regulation, in this case, is expressed in STL form as $\phi^{pedes} = \G[\big(p \wedge (x-x_{cross}\leq 0) \big) \implies \big(x-x_{cross}\leq 0 \ \mathcal{U} \ \lnot p\big) \wedge (v \leq v_{lim})]$ where $x$, $v$ represent position and velocity signals, respectively. Boolean signal $p$ indicates the presence of a pedestrian,  and $v_{lim}$ and $x_{cross}$ are constants denoting the speed limit and the crosswalk position, respectively.
	
\begin{figure*}[!hbt]
		\begin{subfigure}[t]{\textwidth}
			\centering
			\includegraphics[width=\textwidth]{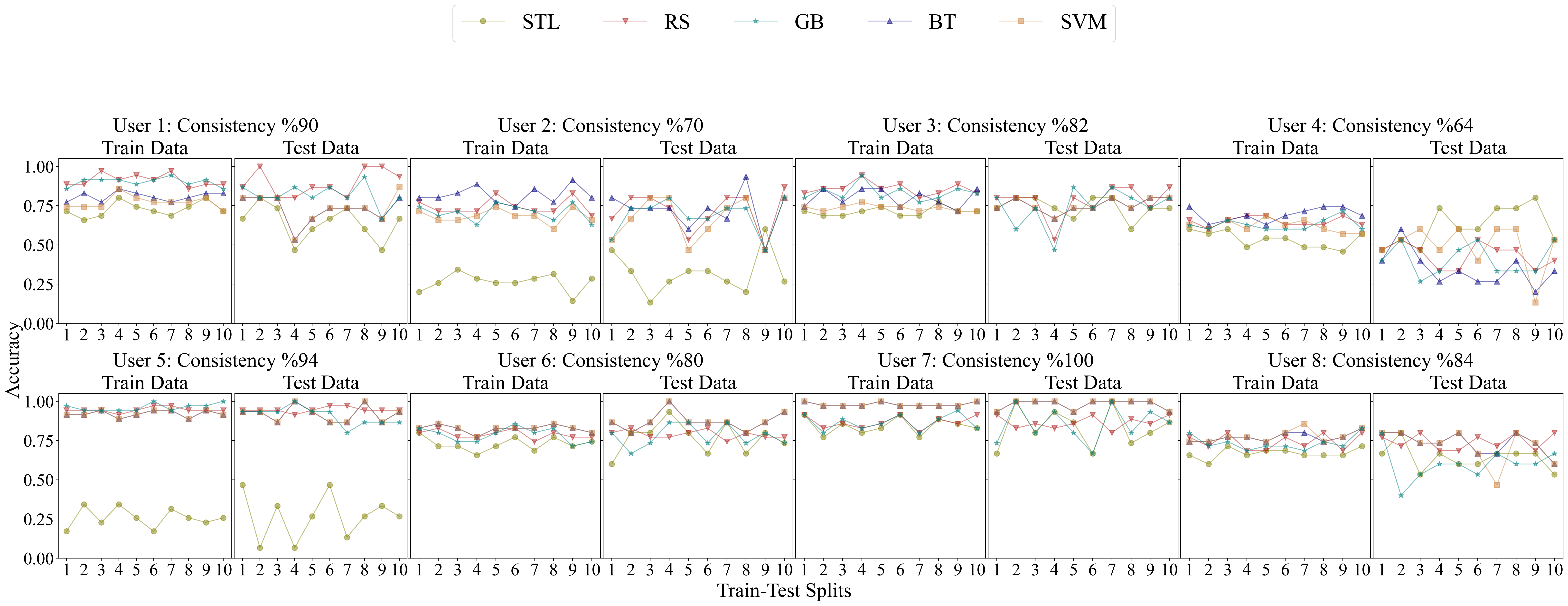}
			\caption{Intersection with a stop sign scenario}
			\label{fig:stop_sign_results}
		\end{subfigure}
		\begin{subfigure}[t]{\textwidth}
			\centering
			\includegraphics[width=\textwidth]{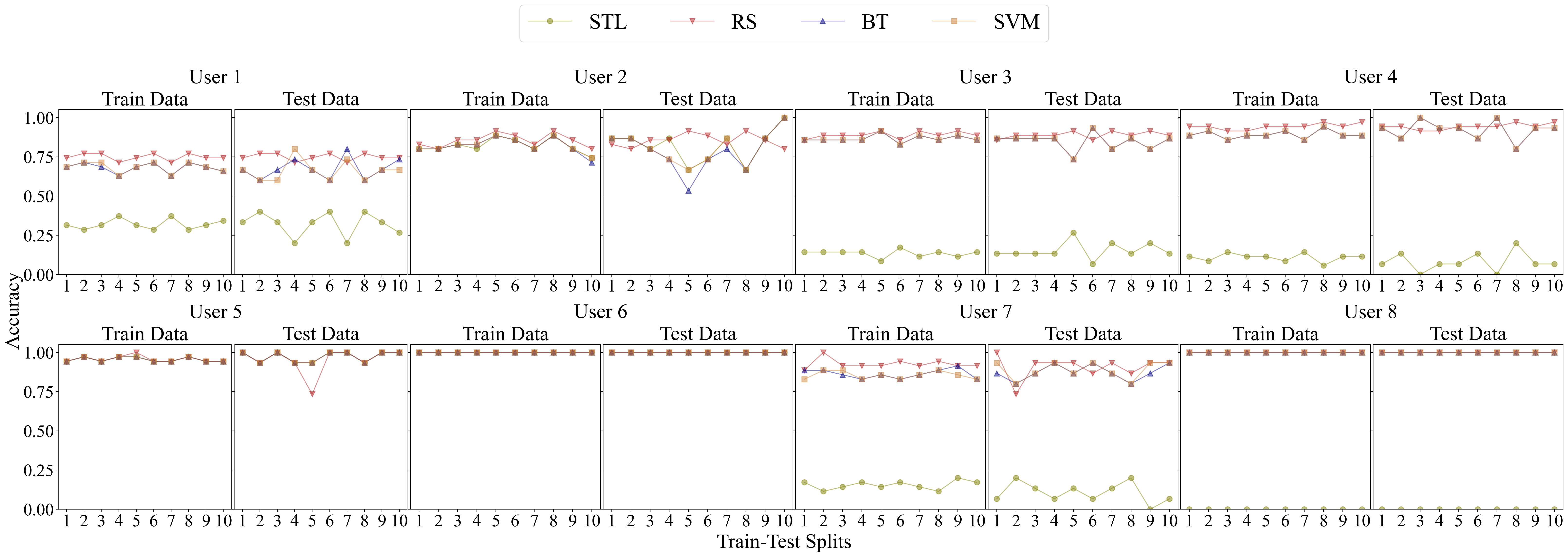}
			\caption{Approaching to a pedestrian scenario}
			\label{fig:pedestrian_results}
		\end{subfigure}
		\caption{Human subject study results for the two scenarios for four of the users. ``STL" denotes the traditional (unweighted) robustness when it is used directly, ``RS" denotes our method with random sampling, ``GB" denotes our method with gradient-based optimization, ``BT" denotes SGD with Bradley-Terry model, and ``SVM" represents SVM classification.}
	\end{figure*}

	\emph{Human Subject Studies:} Studies are completed under IRB Study No. HUM00221976. For each scenario, we collaborate with eight participants with a $75-25\%$ male-female ratio from the $25-35$ age group. We simulate a hundred trajectories that satisfy the temporal logic formula per scenario. We compose fifty pairs such that the Euclidean distance between each pair is greater than a threshold. This threshold value is determined manually as the point at which the difference between signals becomes difficult to discern. These pairs are shown to participants who then choose their preferred behavior. As human decisions can vary in consistency, for the first scenario, we repeat the same question set twice to get a measure of the participant's decisiveness. The consistency levels of four participants are reported in Figure~\ref{fig:stop_sign_results}. 
	
	\subsection{Baseline Methods}\label{sec:benchmark_methods}
	One well-known approach to pairwise preference learning problem is to recast it as a supervised learning problem \cite{Aiolli2011}. Let $\psi(s)$ be the feature vector of item $s$. We construct $\psi(\cdot)$ by dividing the Fourier transform of $s$ into five frequency bins and adding the traditional robustness metric as the final feature. To set up the supervised learning problem, for a given preference pair $(s_i^+, s_i^-)$, we construct a new feature vector as the difference of feature vectors as  $\psi(s_{i}^+) - \psi(s_{i}^-)$. All signal pairs in $\mathcal{P}$ belong to Class $0$. We generate the data for Class $1$ by reversing the signal order and defining the feature vector $\psi(s_i^-)-\psi(s_i^+)$. This process gives us binary labels for all comparison pairs and their reverse orders. Then, we use Support Vector Machines (SVM), with a radial basis function kernel, to learn a binary classifier. For a test pair $(s_1, s_2)$, if $\psi(s_1)-\psi(s_2)$ is classified in Class $0$, we say $s_1$ is preferred over $s_2$; and we say $s_2$ is preferred over $s_1$ otherwise. 
	
	The second baseline method is based on a representation of pairwise user preferences with the likelihood of selecting one item over another. In particular, the Bradley-Terry model is a common likelihood function model in preference learning applications \cite{BradleyTerry}. The Bradley-Terry model\cite{BradleyTerry} uses the following likelihood function:
	\[ P_v(s_{i}^+, s_{i}^-) = \frac{e^{<v,\psi(s_{i}^+)>}}{e^{<v,\psi(s_{i}^-)>}+e^{<v,\psi(s_i^-)>}},\]
	where $\psi(\cdot)$ again represents the feature vector described earlier. Then, we solve for weights $v$ to maximize the log-likelihood as follows:
	\begin{equation}\label{eq:bradley_terry}
		v^* \in \arg \min - \sum_{i=1}^{P}{\log(P_v(s_i^+, s_i^-))}.
	\end{equation}
	In particular, we use stochastic gradient descent (SGD) for solving this problem. Finally, for a test pair $(s_1, s_2)$, if $e^{<v^*,\psi(s_1)>} > e^{<v^*,\psi(s_2)>}$, we say $s_1$ is preferred over $s_2$; and we say $s_2$ is preferred over $s_1$ otherwise.
	
	\subsection{Comparison of Solution Approaches}
	
	In this section, we compare the performance of the proposed solution approaches with the baseline methods listed in Section~\ref{sec:benchmark_methods}. We use the percentage of training (test) pairs that a model accurately predicts as the metric for comparison. 
	
	For each participant, we use ten random $70\%-30\%$ splits of the preference set in to train-test data, i.e., $35$ pairs for the training set and $15$ for the test set. For each split, we compute the train-test accuracy with respect to traditional STL and compare two proposed approaches with two baseline methods. Our first method solves Problem~\ref{prob:minprob} using $\phi^{stop}$ and $\phi^{pedes}$ for respective scenarios, via random sampling with a threshold condition, where we sample $1000$ weight valuations per split. For the stop sign scenario, our second method solves Problem~\ref{prob:minprob} with gradient-based optimization over the loss function $\mathcal{L}$, initialized eleven times, ten from random weight valuations and one from the traditional STL valuation. We report the best training/test accuracy pair among these 11 as a result. The learning rate is $10^{-5}$, $\epsilon = 0.01$, and $\theta = 0.01$. The softness coefficient for \texttt{softmax} is $\beta = 10^{10}$. We terminate the optimization when the cost value difference drops below $10^{-6}$. We divide the training set into batches of five pairs.  Batch selection is random at each iteration. The third method is the SVM classification baseline. For the last method, we solve Equation~\eqref{eq:bradley_terry} via SGD with the learning rate of $0.1$. 
	
	Some representative results are shown in Figures~\ref{fig:stop_sign_results} and \ref{fig:pedestrian_results}. The average performance of all methods for all users and splits is shown in Table~\ref{tab:average_results}. Random sampling gives competitive results with the baseline methods in the stop sign scenario, and outperforms all methods in the pedestrian. Although the gradient-based method gives equally good results as random sampling for the stop sign scenario, it is much slower. Indeed, it was too slow to converge for the pedestrian scenario, due to formula complexity, and is skipped. We conclude that simple random sampling effectively identifies promising weight valuations that improve the traditional STL accuracy, and give comparable results to other methods.
	
	\begin{table}[h!]
		\setlength\tabcolsep{1.5pt}
		\caption{Average accuracy results for different methods on human subject studies. Values represent the average accuracy over all splits and all eight users for each method.}
		\centering
		\begin{tabular}{ccccccccc}
			\toprule
			Method   & \multicolumn{2}{c}{RS (ours)} & \multicolumn{2}{c}{GB (ours)} & \multicolumn{2}{c}{BT}  &  \multicolumn{2}{c}{SVM}   \\ 
			Accuracy & Train & Test & Train & Test & Train & Test & Train & Test\\ \midrule
			Stop sign & $81.2\%$ & $\boldsymbol{77.0\%}$ & $80.2\%$ & $72.4\%$& $\boldsymbol{82.7\%}$& $75.7\%$ & $78.9\%$ & $76.7\%$ \\
			Pedestrian & $\boldsymbol{91.5\%}$ & $\boldsymbol{91.4\%}$ & N/A & N/A & $88.4\%$ & $88.4\%$ & $88.3\%$ & $88.7\%$ \\
			\bottomrule
		\end{tabular}
		\label{tab:average_results}
	\end{table}
	
	Finally, when we look at Figure~\ref{fig:stop_sign_results}, we see that with decreasing consistency, the generalizability of all methods decreases, i.e., they perform poorly on test data.
	
	Now we turn our attention to the safety of different approaches. Ideally, when presented with a pair of signals where one is violating the traffic rules and the other is satisfying, an approach should give preference to the satisfying one. Our method satisfies this nice property by construction. To test how the baselines do in this case, we simulate a hundred violating pairs for the intersection with a stop sign scenario and pair them with satisfying signals. Now, we have fifty satisfying-satisfying signal pairs that we use in human subject studies and a hundred satisfying-violating signal pairs. We create two different training sets: (i) one with fifty satisfying-satisfying pairs, and (ii) one with fifty satisfying-violating pairs in addition to satisfying-satisfying pairs. Test sets are hundred satisfying-violating pairs, and fifty satisfying-violating pairs, respectively. Table~\ref{tab:combined_results} shows the safety performance of two baseline methods and random sampling for the stop sign scenario and one participant. As we can see, baseline methods trained with only satisfying signals perform poorly when encountered with violating signals. However, it is not always feasible to generate real-life behaviors that violate a rule for safety-critical scenarios. When training baseline methods, we rely on simulators to generate violating signals, which may not be realistic. When we look at the results with training set (ii), the test performance of both baseline methods increases, and SVM reaches $100\%$ accuracy. However, none of the baseline methods ensure the safety of arbitrary safe-unsafe pairs.
	
	\begin{table}[h!]
		\setlength\tabcolsep{4.2pt}
		\caption{Safety-critical selection comparison with baseline methods. The test values indicate the percentage of test cases for which the learned model prefers a rule-following (safe) behavior to a rule-violating (unsafe) one.}
		\centering
		\begin{tabular}{ccccccc}
			\toprule
			Method &  \multicolumn{2}{c}{RS (ours)} &  \multicolumn{2}{c}{BT}  &  \multicolumn{2}{c}{SVM}  \\ 
			Trained with & (i) & (ii) & (i) & (ii) & (i) & (ii) \\ \midrule
			Training Accuracy & $92\%$  & $96\%$  & $78\%$ & $82\%$ & $76\%$ & $85.33\%$ \\
			Test Accuracy     & $100\%$ & $100\%$ & $40\%$ & $87\%$ & $31\%$ & $\boldsymbol{100\%}$ \\
			\bottomrule
		\end{tabular}
		\label{tab:combined_results}
	\end{table}
	
	\section{Conclusion, Limitations, and Future Work}
	This work introduced a safe preference learning approach and evaluated its performance in two different driving scenarios.  
	Considering three desirable properties of preference learning for safe personalization mentioned in the introduction, our results show that our method gives competitive results with the baselines in terms of expressivity but significantly outperforms them in terms of safety. Moreover, it is not clear how models learned by generic preference learning methods can be used in control design, whereas our STL-based method can be readily integrated into control synthesis. 
	
	We note that neither random sampling nor gradient-based method guarantees finding an optimal value. We also observe the gradient-based method to have difficulties in convergence for certain formulas. It would be interesting to study different smooth robustness metrics to see if they can mitigate this issue. While preference data in our experiments appears to be on a smaller scale, expecting humans to select preferences for hundreds of signal pairs all at once is impractical. Our experience shows that even dealing with fifty pairs could be overwhelming. To this end, our upcoming focus is on an active learning scheme that maximizes inference using a minimum amount of question pairs. Additionally, we aim to integrate the final WSTL formula into a downstream control synthesis algorithm as in \cite{Cardona2023} to demonstrate its use in control design and to run further validation studies.
	
	\balance

	\bibliographystyle{IEEEtran}
	\bibliography{references.bib}	
	
\end{document}